\documentclass{article}
\usepackage{ijcai25}

\usepackage{times}
\usepackage{soul}
\usepackage{url}
\usepackage[hidelinks]{hyperref}
\usepackage[utf8]{inputenc}
\usepackage[small]{caption}
\usepackage{graphicx}
\usepackage{amsmath}
\usepackage{booktabs}
\usepackage[table]{xcolor}
\usepackage{colortbl}
\usepackage{diagbox} % 用于制作左上角斜杠
\usepackage{multirow} 
\usepackage{url}
\usepackage{booktabs}
\usepackage{amssymb}
\usepackage{bbding}
\usepackage{pifont}
\usepackage{wasysym}
\usepackage{utfsym}
\usepackage{fontawesome}
\usepackage{amsthm}
\usepackage{algorithm}
\usepackage{algorithmic}
\usepackage[switch]{lineno}
\usepackage[T1]{fontenc}

\usepackage{color}
\newcommand{\fzl}[1]{{\color{black}#1}}

\title{%\emph{CorrDetail}: 
\emph{CorrDetail:}
Visual Detail Enhanced Self-Correction for Face Forgery Detection
}

\author{
Binjia Zhou$^{1}$ \footnotemark[2]\and
Hengrui Lou$^{1}$ \footnotemark[2]\and
Lizhe Chen$^2$\And
Haoyuan Li$^3$\and
Dawei Luo$^4$\and
Shuai Chen$^4$\and
Jie Lei$^5$\and
Zunlei Feng$^{1}$\and
Yijun Bei$^1$ \footnotemark[1]\\
\affiliations
$^1$School of Software Technology, Zhejiang University\\
$^2$Shenzhen International Graduate School, Tsinghua University\\
$^3$School of Computer Science and Technology, Zhejiang University\\
$^4$Ant Group\\
$^5$School of Computer Science and Technology, Zhejiang University of Technology\\
}

\begin{document}

\maketitle
\renewcommand\thefootnote{} % 清除脚注编号
\footnotetext{* Corresponding author. Email: beiyj@zju.edu.cn}
% \footnotetext[2]{† These authors contributed equally to this work.}

\begin{abstract}
%\fzl{
With the swift progression of image generation technology, the widespread emergence of facial deepfakes poses significant challenges to the field of security, thus amplifying the urgent need for effective deepfake detection.
Existing techniques for face forgery detection can broadly be categorized into two primary groups: visual-based methods and multimodal approaches. The former often lacks clear explanations for forgery details, while the latter, which merges visual and linguistic modalities, is more prone to the issue of hallucinations. 
To address these shortcomings, we introduce a visual detail enhanced self-correction framework, designated  \emph{CorrDetail}, for interpretable face forgery detection. \emph{CorrDetail} is meticulously designed to rectify authentic forgery details when provided with error-guided questioning, with the aim of fostering the ability to uncover forgery details rather than yielding hallucinated responses. Additionally, to bolster the reliability of its findings, a visual fine-grained detail enhancement module is incorporated, supplying \emph{CorrDetail} with more precise visual forgery details. Ultimately, a fusion decision strategy is devised to further augment the model's discriminative capacity in handling extreme samples, through the integration of visual information compensation and model bias reduction. 
Experimental results demonstrate that \emph{CorrDetail} not only achieves state-of-the-art performance compared to the latest methodologies but also excels in accurately identifying forged details, all while exhibiting robust generalization capabilities.
%}

%With the continuous advancement of image generation technology, face deepfakes pose significant threats to the security domain, thereby increasing the urgency of deepfake detection. Existing face Forgery detection methods can be broadly divided into two categories: visual-based and multimodal techniques. The former lacks visualization of forgery clues, making it resemble a black box and difficult to regulate, while the latter, which combines multiple modalities, is more prone to hallucination issues that can lead to misjudgments. To address these challenges, we propose a multimodal framework based on reverse question-answering and detail enhancement, called \emph{CorrDetail}. Through error-guided questioning, the model is compelled to extract maximum image information to complete calibration answer. Furthermore, to ensure the reliability of the results, we implement fine-grained detail enhancement, enabling the model to extract details more accurately and effectively. Finally, a decision enhancement method based on visual information compensation and model bias mitigation further strengthens the model's discriminative ability on extreme samples. Experiments demonstrate that our method not only achieves state-of-the-art (SOTA) performance among the latest approaches but also accurately identifies forged details and possesses strong generalization capabilities.
\end{abstract}

\section{Introduction}
The rapid advancement of generative technologies, exemplified by Artificial Intelligence Generated Content (AIGC)~\cite{r1}, has profoundly liberated societal productivity, enabling innovative applications across diverse sectors such as news, literature, entertainment, education, and industry. However, the capability of AIGC to produce high-quality content from simple prompts and guidance has simultaneously lowered the barriers for malicious actors to fabricate deceptive information~\cite{r2}. This ease of content falsification has introduced significant challenges in verifying identity information, resulting in an increase in financial fraud, misinformation dissemination, and identity theft, thereby posing severe threats to societal security.
\begin{figure}[!t]
    \centering
    \includegraphics[width=0.47\textwidth]{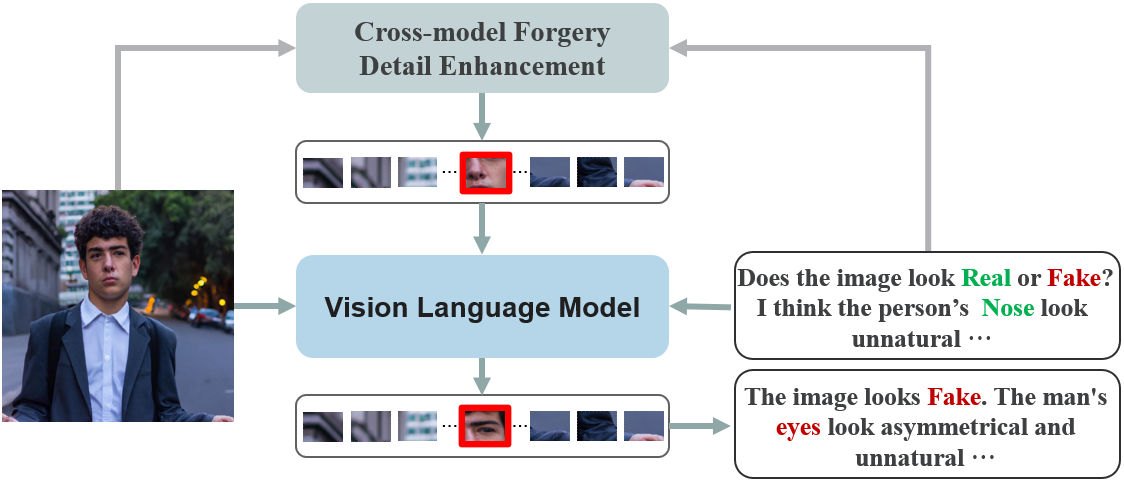}
    \caption{
    %\fzl{
    The illustration for \emph{CorrDetail}, which is designed to correct authentic forgery details  ("eyes look asymmetrical...") when provided with error-guided questioning ("nose look unnatural...") with the assistance of visual fine-grained detail enhancement. 
    %}
    %Our proposed TEVQA method informs the model of the correct or incorrect solutions during the questioning phase, guiding the model to produce accurate answers.
    }
    \label{fig:TEVQA}
\end{figure}

To address the security threats posed by the  facial image forgery technologies, researchers have focused on identifying visual artifacts inherent in forged facial images by analyzing the images themselves. Current approaches~\cite{57,58,59,60,64} involve forensic detection from various perspectives, including the network architectures of forgery methods (such as GANs, autoregressive models, and diffusion models) and the domain-specific features of these methods (encompassing both frequency and spatial domains). These techniques have demonstrated commendable performance in facial authentication tasks. However, concentrating solely on forgery detection within a single visual modality  encapsulates the forgery process as a black box. Models that rely on a single visual modality typically yield only binary forgery classification results, lacking the provision of critical visual cues necessary for understanding the basis of the discrimination. Consequently, this leads to a significant deficiency in the interpretability of the forgery detection process.

To enhance the interpretability and informational depth of facial image forgery detection models, researchers have endeavored to incorporate textual information, thereby developing multimodal deepfake detection frameworks. Leveraging established text-image encoding technologies such as CLIP~\cite{clip} and BLIP~\cite{blip}, a bridge for feature interaction between the image modality and the text modality has been constructed. Within the field, two primary approaches have emerged: text-image information interaction and the application of large-scale visual language models (VLMs). The first approach involves integrating corresponding textual information with forged facial image data to augment the dimensionality of feature extraction for forgery detection.  %This includes techniques such as the fusion of extracted text and image features and contrastive learning between text and image modalities~\cite{wt1,wt2}.  
The second approach builds upon this foundation by replacing traditional neural network models with more extensive and comprehensively pre-trained VLMs.  Through a question-and-answer dialogue mechanism, VLMs can provide detailed semantic information during the deepfake detection process, including specifics about the forged regions, methods employed, and the overall forgery framework~\cite{vlm1,zhang2025common}.  However, within the domain, the training and inference mechanisms of VLMs are susceptible to hallucination phenomena, which not only degrade model performance but also risk supplying erroneous forgery cues.  This poses significant challenges to the reliability and accuracy of multimodal deepfake detection.

%To address these challenges, we propose a novel framework named Visual Detail Enhanced Self-correction for Face Forgery Detection (\emph{CorrDetail}). By leveraging the reverse question-answering capabilities of VLMs and employing multi-branch information completion across text and image modalities, our approach efficiently trains a universal agent capable of dynamically interacting within complex scenarios while providing accurate forgery cues for facial forgery detection. Additionally, the dual-round VLM reasoning mechanism facilitates bias correction within the agent. Our specific contributions are as follows:%

To address these challenges, we propose a visually enhanced self-correction framework, denoted as \emph{CorrDetail}. By providing error-guided questions, \emph{CorrDetail} establishes a \fzl{corrective} %bidirectional
question-and-answer mechanism that cultivates the ability to identify forged details and rectifies training environments prone to hallucinations. In addition, a visual fine-grained detail enhancement module is integrated to furnish \emph{CorrDetail} with more precise visual forgery cues. Finally, we devise a fused decision strategy that combines visual information compensation with model bias reduction, further strengthening the model’s ability to discriminate extreme samples. Our specific contributions are as follows:

\begin{itemize}
%\fzl{
    \item  \fzl{We propose a multimodal forgery detail self-correction framework, alongside the creation of a Self-Correction Visual Question Answering (SCVQA) dataset tailored for facial forgery detection. This framework seamlessly combines a self-correction strategy with a cross-model forgery detail enhancement module, aimed at identifying authentic forgery characteristics and producing intelligible, explanatory text-based forgery clues.}
    \item  We introduce a novel inference decision mechanism that merges a dual-round question-answering process with multi-scale visual feature interactions, effectively correcting biases inherent in large models. This mechanism entails the construction of multi-expert reasoning decisions, thereby bolstering the reliability and precision of the forgery detection process.
    \item Extensive comparative, ablation, and visualization experiments demonstrate that our approach not only achieves state-of-the-art performance but also delivers precise and accurate forgery details, showcasing remarkable generalization capabilities.
    %}
    % \item We develop a multimodal detail completion module (Cross-modal Semantic Enhancement, CSE) that employs fine-grained feature completion to enhance the interaction between text and image modalities across multiple branches. This enhancement significantly improves the extraction and integration of multimodal features.
    % \item  We implement a novel inference decision mechanism (Decision Fusion) that combines the dual-round question-answering process with multi-scale visual feature interactions to correct biases inherent in large models. This mechanism involves constructing multi-expert reasoning decisions, thereby enhancing the reliability and accuracy of the forgery detection process.
    % \item Extensive comparative, ablation, and visualization experiments demonstrate that our method not only achieves SOTA performance but also provides precise and accurate forgery details, exhibiting exceptional generalization capabilities.
\end{itemize}

\section{Related Works}
\subsection{Face Forgery Technology}

GAN-based and autoregressive-based models were pioneers in the field of image generation. Driven by these models, realistic facial modification techniques such as BigGAN~\cite{BigGAN} and StarGAN~\cite{StarGAN1} emerged, enabling traditional facial forgery methods to be categorized into face swapping, face editing, and face reenactment. With the rapid advancement of AIGC technologies, prompt-guided facial image generation methods have become widely accessible, allowing for the creation of diverse facial images with minimal input, such as textual descriptions or sketches. Notable products like Midjourney~\cite{midjourney2022} and Stable Diffusion~\cite{SD} have surged in popularity by integrating prompts into the AIGC generation process (e.g., diffusion models), thereby facilitating guided and procedural content generation. This prompt-guided AIGC generation approach starkly contrasts with traditional task-oriented methods~\cite{deepfacegen}, significantly increasing the difficulty of developing generalized facial forgery detection models.

\begin{figure*}[!t]
  \centering
  \includegraphics[width=\textwidth]{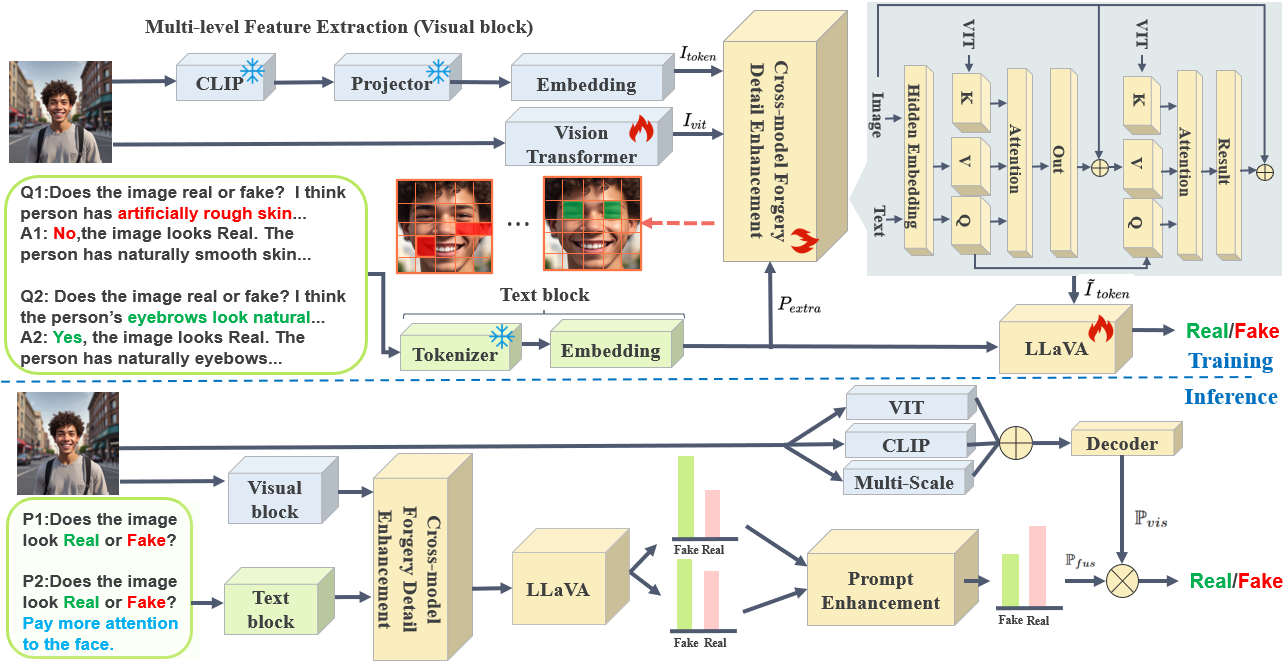}
  \caption{
  \fzl{
  Framework of the proposed \emph{CorrDetail}. During training, the self-correction Q\&A strategy is employed to enhance the forgery clue extraction ability of the VLM, while the visual branch integrates a dual-stream architecture (CLIP + ViT) with the Cross-Model Forgery Detail Enhancement (CFDE) module to provide detailed visual features. In the inference phase, the input image and dual-round prompts are initially processed through the CFDE module for feature alignment and concentration. Subsequently, the final detection outcome is generated through a dual-decision strategy.
  }
  %Framework of the proposed \emph{CorrDetail}, which comprises a training stage and an inference stage. During training, the text modality employs the SCVQA strategy to strengthen reverse reasoning in prompts, while the visual modality adopts a dual-stream architecture (CLIP + ViT) to enhance visual feature extraction. Cross-model Forgery Detail Enhancement (CFDE) module then completes and aligns features across modalities. The Extra-prompt refers to additional facial guidance information (e.g., noise or skin attributes). In the inference stage, the input image and dual-round prompts are first fed into the CFDE module for feature completion and alignment. Subsequently, the final detection result is produced using a dual-decision strategy.
  }
  \label{fig: pipeline}
\end{figure*}

\subsection{Vision-based Face Forgery Detection} 

Vision-based Face Forgery Detection is the predominant approach in the field, concentrating on the forgery artifacts present in manipulated facial images. This methodology can be specifically categorized into frequency-domain-based and spatial-domain-based techniques~\cite{deepfacegen}. The core principle involves amplifying forgery traces through transformations in different domains, thereby capturing clearer and more precise artifacts.

In the frequency domain, forgery features are extracted at a fine-grained level by mapping forged images into high-frequency, mid-frequency, and low-frequency dimensions~\cite{59}. This allows for the detailed identification of subtle manipulation signs that may not be easily discernible in the spatial domain. Conversely, in the spatial domain, the focus is on identifying inconsistencies introduced during the face modification process, such as artifacts, glare, and blurred regions~\cite{color,saturation,blended}. To address the novel generation methods introduced by AIGC, new approaches have emerged that rely on reconstruction errors for forgery detection in the spatial domain. These methods analyze discrepancies between the original and reconstructed images to identify potential manipulations.

While the aforementioned methods effectively accomplish facial image forgery detection, single-modality approaches fail to provide specific semantic information regarding the forgeries. The absence of disclosed forgery cues results in insufficient interpretability of these methods, limiting their ability to offer comprehensive insights into the detection.

\subsection{Multimodal-Based Face Forgery Detection}

Recent advancements in image-text pre-trained models~\cite{llava,clip,blip} have driven researchers to adopt multimodal approaches to enhance face forgery detection. Currently, these approaches can be broadly categorized into contrastive learning-based methods and VLM-based methods.

The contrastive learning-based methods leverage the comparison of encoded image-text features to align and reinforce discriminative features or utilize textual information to amplify forgery traces within the image modality~\cite{wt1,wt2}. In contrast, VLM-based methods utilize VLMs for forgery detection, leveraging their powerful performance for tasks like data processing, feature extraction, and decision-making.

While both categories have demonstrated performance improvements, they exhibit significant areas requiring enhancement. Firstly, current methodologies primarily involve simple encoding and fusion or alignment of multimodal features. This coarse-grained discrimination approach impedes the construction of comprehensive and deeply nuanced features. Furthermore, VLMs are often treated as black-box tools or auxiliary aids within the domain, lacking an in-depth development of feature extraction and discrimination decision models grounded in the fundamental principles of VLMs. More critically, this coarse-grained approach induces hallucination phenomena within VLMs, hindering their ability to deliver specific and accurate discrimination results and analyses. As a result, the models are unable to provide detailed and reliable insights into the nature and extent of the forgeries, significantly compromising the overall effectiveness and trustworthiness of the detection system.

\subsection{Critique and Refine Mechanism }

In the course of in-depth research and application of large language models (LLMs), these models have demonstrated remarkable potential in critique and refine. Researchers have leveraged LLMs to assess the results of specific models, facilitating their self-improvement. In the domain of VLMs, critique and refinement mechanisms have also been introduced to perform logical corrections~\cite{saunders,madaan2023selfrefineiterativerefinementselffeedback,gou2024criticlargelanguagemodels}. Specifically, through  querying of results and multi-round interactions, LLMs are endowed with self-improvement mechanisms that allow them to critique and refine their reasoning capabilities~\cite{LLM1,LLM2,LLM3,LLM4}. Following these advancements, the critique and refine Mechanism has garnered significant attention, and in the benchmark domain, there has been a growing trend to use critique mechanisms to assess the critique-correction reasoning ability of LLMs~\cite{lightman2023letsverifystepstep,li2024generative,luo2024critique,huang2024large}.

The aforementioned work on the critique and refine mechanism in text or multimodal LLMs primarily focuses on correcting logical reasoning capabilities. Inspired by this, we propose a novel method that employs error-agnostic question-guided mechanisms, enabling the model to learn how to identify visual forgery features and provide accurate responses. This method does not focus on correcting logical reasoning abilities but rather aims to prevent hallucinated responses, ensuring more reliable and precise detection of visual forgeries.

\section{Method}
 %Figure~\ref{fig: pipeline} shows the overall framework of \emph{CorrDetail}. In this section, we will introduce the various modules of the proposed method. In Section \ref{sec:TEVQA}, a novel VQA training approach is presented to enhance the learning efficiency of VLM. Section \ref{sec:CSE} proposes a semantic-based attention enhancement method that strengthens the model's focus on specific semantics from the feature perspective. In Section \ref{sec:DF}, we introduce a multi-expert model ensemble decision-making approach, which, through clever handling, allows VLM to controllably pay more attention to facial regions.
 \fzl{
Fig.~\ref{fig: pipeline} shows the overall framework of \emph{CorrDetail}. In this section, we present the various modules of \emph{CorrDetail}. In Section \ref{sec:TEVQA}, we introduce an innovative self-correction VQA training strategy designed to augment the forgery clue extraction capability of the VLM. Section \ref{sec:CSE} unveils the cross-model forgery detail enhancement module, which provides intricate visual features for facial forgery detection. In Section \ref{sec:DF}, we introduce a multi-expert ensemble decision-making strategy, which strengthens the reliability and precision of the final outcome.
 }

\subsection{%Training Enhance Visual Question Answering
Self-Correction Visual Question Answering
}
\label{sec:TEVQA}
In the field of face forgery detection, the VQA training of multimodal large models for related detection tasks is essentially a form of knowledge infusion. Although it can continuously improve model performance through large amounts of data, inspired by~\cite{LLM3} masking strategy and the fact that images dominate the dual-modal interactions with text in the current task, we aim to enable the model to capture the \fzl{forgery visual features} %relevant information 
in images to the greatest extent during training. To this end, we propose a novel training method, as shown in Fig.~\ref{fig:TEVQA}.

Based on the VQA format provided by DD-VQA~\cite{zhang2025common} in the FF++ dataset~\cite{rossler2019faceforensics++}, we enhance the training process by introducing a new prompt in the question. 
%Specifically, during training, we add wrong and authentic forgery detail description to the question. 
\fzl{
During training, we introduce incorrect and authentic forgery details into the question. There is a $70\%$ probability of incorporating erroneous details into the original prompt (Q1 in Fig.~\ref{fig: pipeline} is an example), a $15\%$ probability of adding genuine forgery details through synonym substitution (Q2 in Fig.~\ref{fig: pipeline} is an example), while the remaining questions remain unaltered.
We deliberately refrain from entirely inserting incorrect details, as this could prompt the model to exploit the prompt directly, deriving answers opportunistically.
This Self-Correction Visual Question Answering (SCVQA) dataset equips the VLM to accurately identify authentic forgery details, rather than producing hallucinated responses during the training process.
}
%There is a 70\% probability of making it the antonym, a 15\% probability of synonym substitution, and for the remaining 15\%, we leave the question unchanged. This method is akin to pre-informing the model of a potential solution during training, which might be completely correct or entirely wrong, encouraging the model to develop skepticism and perceptive abilities to extract as much information as possible from the input image. However, we avoid fully revealing the wrong solutions, as this might lead the model to exploit the prompt directly to derive the answer opportunistically. 

%In subsequent experiments, we demonstrated that this training method significantly improves model performance compared to the traditional VQA approach when using the same amount of data.

\subsection{%Cross-modal Semantic Enhancement
Cross-model Forgery Detail Enhancement
}
\label{sec:CSE}
%To further supplement the image details required for counter-questioning correction, we need to guide the model to focus on specific content within the image. However, most methods rely on modifying the prompt. Nonetheless, this attention enhancement method, which is solely based on text input, overly depends on pre-trained knowledge and is not particularly effective in face forgery detection. Considering that large-scale pre-training has already mapped image tokens to the same semantic space as text tokens and that, from the perspective of generation principles, forged images produced by AIGC often exhibit abnormal semantic similarity with text, we aim to enhance the attention of the corresponding modules in images from a semantic feature perspective.
\fzl{
In commonly employed VLM, the intricate visual tokens and succinct text tokens are aligned and mapped into a shared latent space.
During this alignment and mapping process, the sophisticated visual tokens inevitably lose some of their detailed visual features. However, face forgery detection must prioritize the capture of these forged visual nuances.
To address this, we have designed a Cross-model Forgery Detail Enhancement (CFDE) module, which enriches the VLM with additional visual details, enabling it to learn to discern and identify forged visual features.
}

\fzl{
Specifically, as depicted in Fig.~\ref{fig: pipeline}, once the image passes through the CLIP and Projector modules, its embedding is devoid of forgery-specific visual details, such as the intrinsic consistency of surrounding elements and the relational information of relevant positions. Therefore, we introduce an additional vision transformer branch to extract the image's intrinsic visual features, providing supplementary information for face forgery detection.
}

\fzl{
Formally, the extra text prompt $P_{extra}$ is processed through the same tokenizer and fused with the image tokens $I_{token}$ and the image visual feature $I_{vit}$ (processed through the Vision Transformer) in a cross-modal attention mechanism. The extra text prompt is used as the query $Q$, the image tokens as the value $V$, and the image the visual feature as the key $K$ to generate the target attention map. Finally, this attention-enhanced image token $\tilde{I}_{token}$ is obtained by adding the resulting attention map to the original image token in a residual manner. The specific formula is as follows:
}
\begin{equation}
\begin{split}
\tilde{I}_{token}=\mathcal{A}(P_{token},I_{vit},\mathcal{A}(P_{token},I_{vit},I_{token})), \nonumber
\end{split}
\end{equation}
\begin{equation}
\scalebox{1.03}{$
\begin{split}
    P_{token} &= Tokenizer(P_{extra}), \\
    I_{vit} &= VIT(I), \\
    I_{token} &= Projector(Clip(I)), \nonumber
\end{split}
$}
\end{equation}
\begin{equation}
    \mathcal{A}(Q,K,V)=softmax(\frac{QK^{T}}{\sqrt{d_{k}}})V+V, \nonumber
\end{equation}
where $CLIP(\cdot)$ represents LLaVA's own visual encoder, and $Projector(\cdot)$ represents the pre-trained image token mapping module, $\mathcal{A}(Q,K,V)$ denotes the self-attention operation in transformer. The final obtained $\tilde{I}_{token}$ will serve as the new image token input to the VLM.

\begin{figure}[!t]
    \centering
    \includegraphics[width=0.49\textwidth]{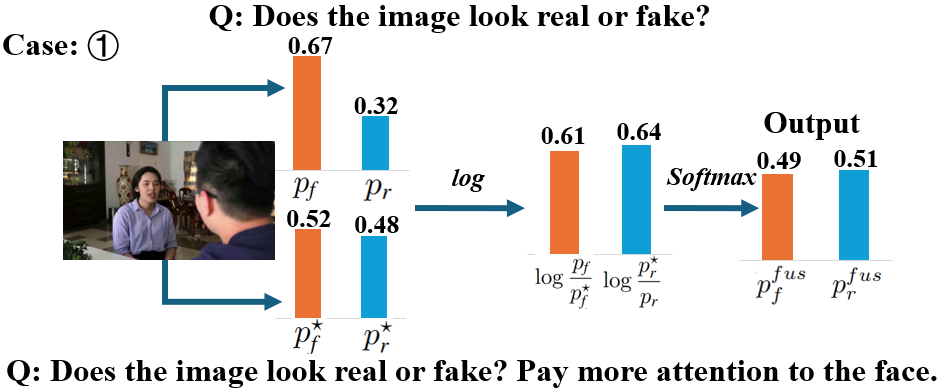}
    \caption{
    \fzl{
    One case for Prompt Enhancement Mechanism handling hard samples (e.g., those with smaller facial proportions). 
    }
    %The illustration of pre- and post-processing on VLM outputs demonstrates that certain challenging positive and negative samples (e.g., those with smaller facial proportions).
    }
    \label{fig:decision fuse}
\end{figure}

\subsection{Decision Fusion}
\label{sec:DF}

\begin{table*}[!t]
\centering
\renewcommand{\arraystretch}{1.2}
\setlength{\tabcolsep}{11pt}
\resizebox{\textwidth}{!}{
\begin{tabular}{c|cc|cc|ccc}
\hline
\multirow{3}{*}{\diagbox[width=14em, font=\Large\centering]{Methods}{Datasets}} & \multicolumn{4}{c|}{Intra-Testing} & \multicolumn{3}{c}{Cross-Testing}\\ \cline{2-8}
& \multicolumn{2}{c|}{FF++ (LQ)} & \multicolumn{2}{c|}{FF++ (HQ)} & \multicolumn{3}{c}{CelebDF (trained on FF++ (LQ))} \\\cline{2-8}
                         & ACC↑            & AUC↑            & ACC↑            & AUC↑     & ACC↑            & AUC↑            & EER↓           \\ \hline
XceptionNet ~\cite{rossler2019faceforensics++}     & 86.86           & 93.50            & 95.04           & 96.30            & 60.11           & 61.80            & 41.73           \\
EN-B4 ~\cite{tan2019efficientnet}           & 86.67           & 88.20            & 96.63           & 99.18           & 66.63           & 67.22           & 38.14           \\
Local-relation ~\cite{chen2021local}    & 91.47           & 95.21           & 97.59           & 99.56  & \multicolumn{3}{c}{--} \\ 
RFM ~\cite{wang2021representative}             & 87.06           & 89.83           & 95.69           & 98.79           & 64.42           & 65.64           & 38.80            \\
RECCE ~\cite{cao2022end}             & 91.03           & 95.02           & 97.06           & 99.32           & 67.96           & 68.71           & 35.73  \\ 
CD-Net ~\cite{song2022adaptive}            & 88.12           & 95.20            & 98.75           & 99.90            & \multicolumn{3}{c}{--} \\    
UIA-VIT ~\cite{zhuang2022uia}         & 91.05           & 94.88           & 98.62           & 99.33           & \underline{69.80}   & \underline{70.15}  & 35.82           \\
GS ~\cite{guo2023controllable} & 92.76  & \underline{96.85}  & \underline{99.24}   & \textbf{99.95}   & \multicolumn{3}{c}{--} \\
HiFi-Net ~\cite{guo2023hierarchical}        & 89.25           & 92.10            & 97.41           & 98.56           & 67.20            & 68.80            & 36.13           \\
PFG-DD ~\cite{lin2024preserving}    & 91.03           & 93.47           & 97.33           & 98.68           & 68.51           & 69.68           & 35.94         \\
RECCE (DD-VQA) ~\cite{zhang2025common}  & \underline{92.08}  & 95.36           & 98.65           & 99.79         & 69.46           & 70.21           & \underline{35.63}   \\ \hline
\emph{CorrDetail} (Ours)& \textbf{94.41} & \textbf{96.93}  & \textbf{99.28}   & \underline{99.92}  & \textbf{72.32}   & \textbf{72.80}    & \textbf{33.87}           \\ \hline
\end{tabular}
}
\caption{
%Comparison of methods across datasets with highlighted Best (\textbf{Red}) and Second Best (\underline{Blue}) results, and cross-testing is trained on FF++(LQ). ACC and AUC (↑) indicate higher values are better, while EER (↓) indicates lower values are better. All scores are in \%.
\fzl{
Comparison of \emph{CorrDetail} with SOTA methods across both intra- and cross-datasets. The highest and second-highest performances are highlighted in \textbf{best}  and \underline{second best}, respectively. ACC and AUC (↑) denote that higher values are preferred, while EER (↓) signifies that lower values are desired. All scores are expressed in \%.
}
}
\label{tab:comparison}
\end{table*}

%Even though the SCVQA and CFDE modules have made efforts to enhance and capture image details, facial information in some samples may occupy only a very small portion of the entire image, as shown in Fig.~\ref{fig:decision fuse}. Additionally, because the model makes authenticity judgments based on the whole image, there is a tendency to favor authenticity in cases where the background is not forged using traditional methods, which is unfair for the final result determination.  \fzl{Therefore, we propose a decision enhancement module to mitigate the misjudgments caused by the aforementioned issue.}

%Moreover, the inherent image token mapping in VLM leads to information loss. Therefore, we propose a decision enhancement module to mitigate the misjudgments caused by the aforementioned issues.

%This module is mainly divided into two parts. The first part is the Prompt Enhancement Mechanism, which primarily explains how to post-process the VLM to free the output from the biases and fairness issues caused 
%by the Tendentious misjudgment of face with small porporation images.
%by the overall tendencies of the image. 
%The second part is the Dual-Expert Model Decision, where a separate visual branch is trained independently to mitigate the information loss caused by image embedding in the VLM. Through joint decision-making, the two branches complement each other, achieving stronger performance.

\fzl{
This module is primarily subdivided into two segments. The first segment is the Prompt Enhancement Mechanism, which is chiefly conceived to post-process the VLM, thereby liberating the output from the biases and fairness challenges stemming from the tendentious misjudgment of faces in images with small proportions.
The second segment is the Dual-Expert Model Decision, in which an independent visual branch is trained separately to alleviate the information loss induced by image embedding within the VLM. Through collaborative decision-making, the two branches harmonize, resulting in superior performance.
}

\textbf{Prompt Enhancement Mechanism.} During the testing phase, the same image is provided with a basic question and an additional guiding prompt, such as \textit{"Does the image look real or fake?"} and \textit{"Does the image look real or fake? Pay more attention to the face/background."}  Variations in the text input will inevitably lead to changes in the model's binary classification probability output. 

Therefore, the Prompt Enhancement Mechanism is designed to address the issue where the detection probability diminishes when the model places excessive focus on the face or background with the added guiding prompt, suggesting that the initial detection probability is dubious. Fig.~\ref{fig:decision fuse} illustrates the scenario in which the model's predicted probability for "fake" decreases as more attention is directed towards the face, indicating that the initial "fake" prediction is unreliable. A possible explanation is that the model's initial prediction relies on the background features for challenging samples with smaller facial proportions.

For the initial prediction probability $[p_{f}, p_{r}]$ and second-round prediction $[ p^{\star}_{f},  p^{\star}_{r}]$ with additional guiding prompt, the final fusion prediction $\mathbb{P}_{fus}$ is adjusted as follows: 
\begin{equation}
\mathbb{P}_{fus} =
\begin{cases}
softmax([
\log \dfrac{p_{\textit{f}}}{ p^{\star}_{\textit{f}}}, \log \dfrac{ p^{\star}_{\textit{r}}}{p_{\textit{r}}}]),\ p_{f}> p_{r}\ \&\  p_{f}> p^{\star}_{f}\    \\
softmax([\log \dfrac{ p^{\star}_{\textit{f}}}{p_{\textit{f}}}, \log \dfrac{p_{\textit{r}}}{ p^{\star}_{\textit{r}}}]),\ p_{r}>p_{f}\ \&\ p_{r}> p^{\star}_{r}\   \nonumber
\end{cases}
\end{equation}
when the initial "fake" prediction $p_{f}$ satisfy the condition $(1-\lambda) \geq p_{f} \geq \lambda$, where $\lambda$ is set as a hyperparameter, serving as the exclusion threshold. For $p_{f} > (1-\lambda)$, $p_{f} <\lambda$ or any other cases, final fusion prediction $\mathbb{P}_{fus}$ will simply match the initial prediction $[p_{f},p_{r}]$.

\textbf{Dual-Expert Model Decision.} 
\fzl{
As depicted in Fig.~\ref{fig: pipeline}, an additional purely visual branch is trained independently using cross-entropy loss, enabling it to perform forgery detection autonomously. Its output is subsequently integrated with the VLM branch's output, enhancing the probability for discriminative consistency. The specific formula is as follows:
}
%This branch is trained separately using cross-entropy loss and is capable of performing forgery detection independently. The overall structural diagram is presented in Fig.~\ref{fig: pipeline}, and its output is ultimately fused with the output of the VLM branch. The specific formula is as follows:
\begin{equation}
\begin{split}
        \mathbb{P}_{final} &=softmax([p^{fus}_{f} \times p^{vis}_{f},p^{fus}_{r} \times p^{vis}_{r}]),  \nonumber   \\
    \mathbb{P}_{fus}&=[p^{fus}_{f},p^{fus}_{r}],  \mathbb{P}_{vis}=[p^{vis}_{f},p^{vis}_{r}],  \nonumber
\end{split}
\end{equation}
where $\mathbb{P}_{fus}$ represents the output of the VLM branch, $\mathbb{P}_{vis}$ represents the output of the traditional visual branch, and finally $\mathbb{P}_{final}$ is the final discriminative output.

\begin{figure}[!t]
    \centering
    \includegraphics[width=0.47\textwidth]{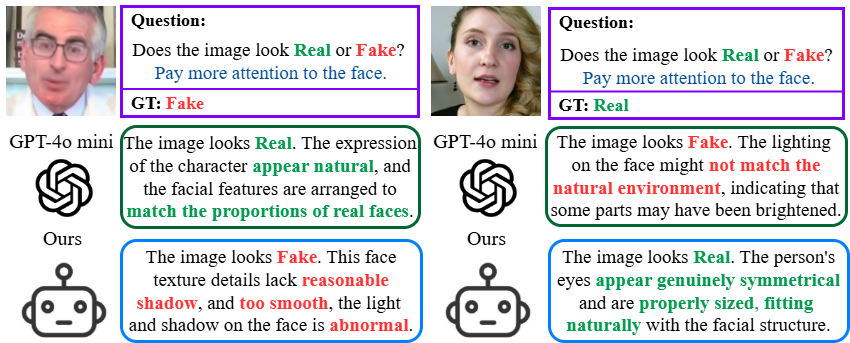}
    \caption{
    %Qualitative examples: The visualization of results on certain samples trained using the TEVQA approach. For failure cases of GPT-4o mini, our method achieves correct classification results along with corresponding explanations.
    \fzl{
    Qualitative examples of results on certain samples. For failure cases of GPT-4o mini, \emph{CorrDetail} achieves correct classification results along with corresponding explanations.
    }
    }
    \label{fig:example1}
\end{figure}

\begin{figure*}[!t]
    \centering
    \includegraphics[width=0.88\textwidth]{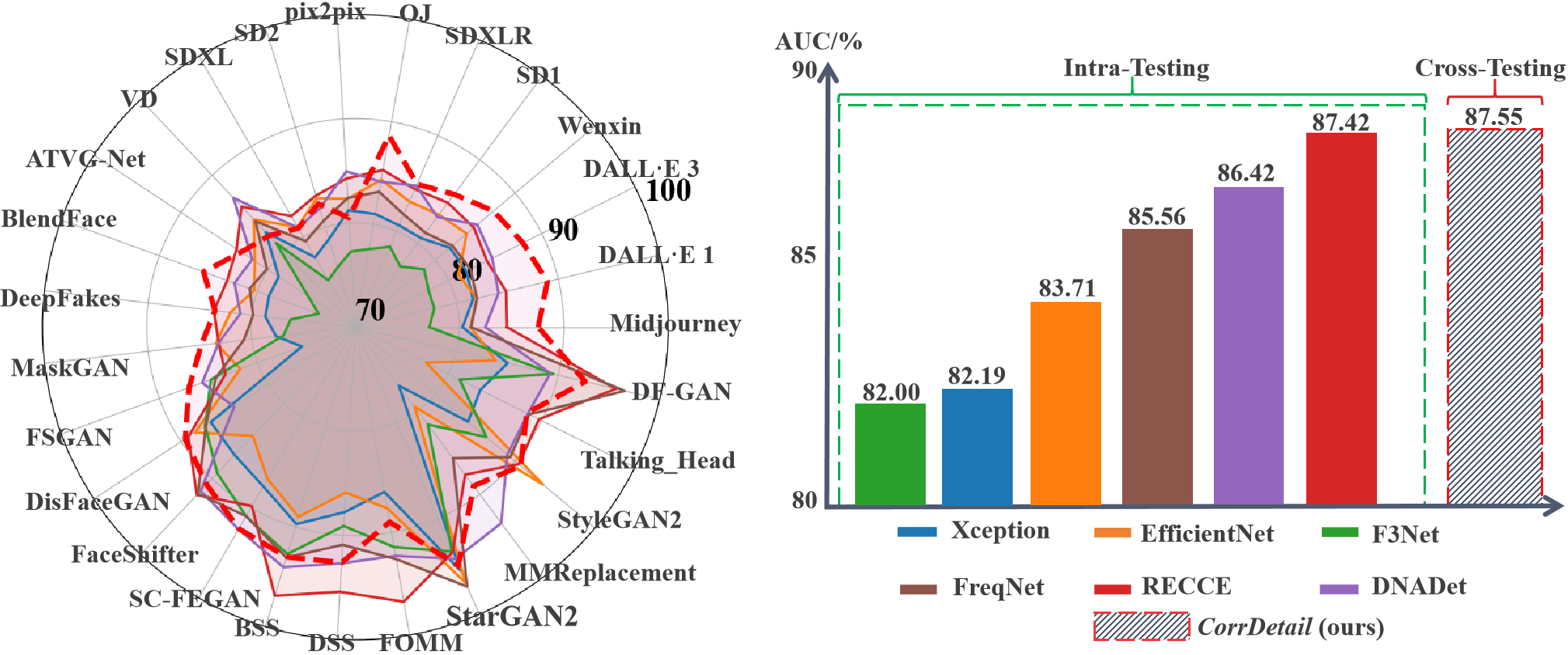}
    \caption{
    \fzl{
The generalization performance of \emph{CorrDetail} in comparison to the intra-performance of existing methods. The left radar chart illustrates the detection efficacy across various types of forgery samples, with solid lines representing the comparison techniques and a dashed line denoting \emph{CorrDetail}. The bar chart on the right displays the average AUC for each method. The six comparison methods are trained and evaluated on the DeepFaceGen dataset. In contrast, \emph{CorrDetail} is trained on SCVQA and tested on DeepFaceGen.
    }
    %Performance comparisons on Deepfacegen (AUC, \%). The radar chart on the left illustrates the detection performance of \emph{CorrDetail} compared to 6 other methods, with solid lines representing the comparison methods and a dashed line representing \emph{CorrDetail}. The bar chart on the right displays the average AUC for each method. The 6 comparison algorithms enclosed within green dashed boxes represent Intra-testing, meaning they were both trained and tested on the DeepFaceGen dataset. In contrast, \emph{CorrDetail}, highlighted within a red dashed box, was trained on DD-VQA (VQA pairs constructed from the FF++ dataset) and tested on DeepFaceGen.
    }
    \label{fig:leida}
\end{figure*}

\section{Experiments}
In this section, we introduce the experimental setup and environment, present a performance comparison between \emph{CorrDetail} and recent approaches across various datasets, and highlight the advantages of our approach. Finally, we conduct ablation studies to further validate the effectiveness of each module.

\textbf{Architecture and Implementation Details.}
The vision branch in \ref{sec:DF} consists of three components: CLIP, VIT, and Multi-scale. The CLIP module corresponds to the corresponding part in the VLM's architecture, VIT primarily uses vit-L/16, and Multi-scale extracts features at different scales by progressively decomposing through multi-level sampling. More details on the structure of each module can be found in the supplementary materials.
We fine-tuned LLaVA-1.5-7B on the SCVQA dataset using a learning rate of $5e^{-5}$, training for $3$ epochs with a batch size of $64$ on three Nvidia A100 GPUs. The training process employed cosine annealing learning rate decay and the AdamW optimizer. The threshold $\lambda$ is set as $0.1$. More details are given in the \emph{supplementary materials}.

\textbf{SCVQA Dataset.} Based on the original data from the FF++ dataset, we have constructed the SCVQA dataset, which consists of 19,797 [image, question, answer] triples to support the self-correction mechanism. The dataset includes three types of data formats: (1) 70\% of the data is in the form of [image, question, incorrect detail description as the answer], (2) 15\% is in the form of [image, question, null answer], and (3) 15\% is in the form of [image, question, correct detail description as the answer].

\begin{table}[!t]
\centering
\renewcommand{\arraystretch}{1.3} % 调整行高
\setlength{\tabcolsep}{2.5pt} % 调整列间距
\resizebox{0.45\textwidth}{!}{
\begin{tabular}{c|cc|cc}
\hline
\multirow{2}{*}{\diagbox[width=10em,font=\Large\centering]{Methods}{Datasets}} 
    & \multicolumn{2}{c|}{DFR} & \multicolumn{2}{c}{WDF} \\ \cline{2-5}
    & AUC↑           & ERR↓           & AUC↑           & ERR↓           \\ \hline
EN-B4 ~\cite{tan2019efficientnet}         & 92.18          & 15.51          & 67.89          & 37.21          \\
Xception ~\cite{rossler2019faceforensics++}      & 91.93          & 15.52          & 68.90           & 38.67          \\
VIT-B ~\cite{dosovitskiy2020image}             & 80.47          & 26.97          & 75.29          & 33.40           \\
LTW ~\cite{sun2021domain}           & \multicolumn{2}{c|}{--}  & 67.12          & 39.22          \\ 
GFF ~\cite{luo2021generalizing}           & \multicolumn{2}{c|}{--} & 66.51          & 41.52          \\
DCL ~\cite{sun2022dual}           & 92.26          & 14.91          & 72.95          & 35.73          \\ 
RECCE ~\cite{cao2022end}         & 92.93          & 14.74          & 74.38          & 32.64          \\ 
CFM ~\cite{luo2023beyond}           & 95.18          & 11.87          & 78.39          & 30.79          \\  
PFG-DD ~\cite{lin2024preserving}         & 94.74          & 12.59          & 77.41          & 29.50          \\
MOE-FFD ~\cite{kong2024moe}           & \underline{95.39} & \underline{11.29} & \underline{80.64} & \underline{27.11}  \\ 
\emph{CorrDetail} (Ours)   & \textbf{96.77}  & \textbf{10.68}  & \textbf{81.58}  & \textbf{26.88}         \\ \hline
\end{tabular}}
\caption{
\fzl{
The generalization performance comparison of \emph{CorrDetail} and other methods trained on FF++ (HQ). Missing data is represented by "--".
}
%Comparison of Methods Across Datasets, all methods are trained on FF++ (HQ). AUC (↑,\%) indicates higher values are better, while ERR (↓,\%) indicates lower values are better. Missing data is represented by `--'.
}
\label{tab:comparison}
\end{table}

\textbf{Evaluation Dataset and Metrics.} We evaluate the proposed method using the FF++~\cite{r12}, CelebDF~\cite{r13}, WildDeepfake (WDF)~\cite{zi2020wilddeepfake}, DeepForensics-1.0 (DFR)~\cite{jiang2020deeperforensics}, and DeepFaceGen~\cite{deepfacegen} datasets. The WDF contains 7,314 facial action sequences extracted from 707 DeepFake videos, and the DFR represents the largest face forgery detection dataset by far, with 60,000 videos constituted by a total of 17.6 million frames. CelebDF employs face swapping as the primary forgery technique, encompassing a total of 5,639 videos. FF++ includes four forgery methods—Deepfake, Face2Face, FaceSwap, and NeuralTextures—comprising a total of 4,000 videos. The videos from both FF++ and CelebDF are cropped into images for facial forgery detection, establishing them as widely adopted evaluation datasets. DeepFaceGen is a facial forgery dataset that includes both video and image modalities. From this dataset, we selected facial forgery images using both traditional task-oriented and novel prompt-guided forgery methods, encompassing a total of 27 approaches. Consistent with the evaluation metrics established in the benchmark papers, we adopt Accuracy (ACC), Equal Error Rate (EER), and Area Under the Curve (AUC) as metrics.

\subsection{Comparison with SOTAs}

%In this section, we adhere to the evaluation methodology outlined in~\cite{zhang2025common}, fine-tune \emph{CorrDetail} on SCVQA, and conduct both intra-domain and cross-domain performance tests on FF++ and CelebDF, respectively.

In this section, We compare \emph{CorrDetail} with several mainstream  methods, including XceptionNet, HiFi-Net, RECCE, CD-Net, Local-relation, RFM, RECCE (DD-VQA), EN-B4, GS, and UIA-VIT. We fine-tune \emph{CorrDetail} on SCVQA constructed with FF++.
As presented in Table \ref{tab:comparison}, \emph{CorrDetail} achieves over 94\% in both ACC and AUC in the intra-testing evaluation on FF++, maintains over 72\% in ACC and AUC in cross-testing on CelebDF, and reduces the EER to below 34\%, attaining  the best performance in six out of seven evaluation metrics across intra-testing and cross-testing. The remaining metric also attains the second-best performance. Notably, \emph{CorrDetail} demonstrates a significant advantage over existing SOTA models in the field, specifically RECCE (DD-VQA), which is based on VLM, and GS, which is based on feature decoupling. Fig.~\ref{fig:example1} further illustrates \emph{CorrDetail}'s ability to accurately detect instances of facial forgery and provide precise forgery cues.

\subsection{Generalization on Different Datasets}

To further validate the effectiveness of \emph{CorrDetail}, we conducted generalization performance experiments. Specifically, on CelebDF, we trained \emph{CorrDetail} on the SCVQA and performed cross-dataset evaluation by testing it on the CelebDF. We compared the performance of \emph{CorrDetail} against other compared methods that were also trained on FF++. On DeepFaceGen, we evaluated \emph{CorrDetail} against  six SOTA detection models. It is important to note that the six detection models were subjected to intra-testing, meaning they were both trained and tested on DeepFaceGen. In contrast, \emph{CorrDetail} was trained on SCVQA and tested on DeepFaceGen, representing a cross-testing scenario.

The experimental results for the CelebDF are illustrated in Table~\ref{tab:comparison}. In the cross-testing experiments presented in the right column, our method achieved the best performance, surpassing all comparison methods. For the more challenging DeepFaceGen benchmark, the results depicted in Fig.~\ref{fig:leida} demonstrate that, even under the less favorable experimental setup where \emph{CorrDetail} was trained on SCVQA and tested on DeepFaceGen while the comparison algorithms were trained and tested on DeepFaceGen, \emph{CorrDetail} still outperformed the SOTA model RECCE.

In these unseen data scenarios, encompassing both traditional forged facial data and novel AIGC-generated forged data, \emph{CorrDetail} consistently delivered outstanding performance. These results conclusively demonstrate the robustness and effectiveness of \emph{CorrDetail} in diverse and challenging forgery detection environments.

\begin{table}[!t]
\centering
\renewcommand{\arraystretch}{1.1} % 调整行高
\setlength{\tabcolsep}{2pt} % 调整列间距
\resizebox{0.48\textwidth}{!}{
\begin{tabular}{ccc|cc|c}
\hline
\multicolumn{3}{c|}{Module} & \multicolumn{2}{c|}{Intra-Testing} & \multicolumn{1}{c}{Cross-Testing}\\
\hline
\multirow{1}{*}{SCVQA} & \multirow{1}{*}{CFDE} & \multirow{1}{*}{Decision Fusion} & \multicolumn{1}{c}{LQ} & \multicolumn{1}{c|}{HQ} & \multicolumn{1}{c}{CelebDF} \\ \hline
\XSolidBrush               & \XSolidBrush           & \XSolidBrush             & 88.44              & 91.87      & 65.58    \\ 
\XSolidBrush              & \XSolidBrush         & \CheckmarkBold             & 90.59               & 93.48    & 67.15       \\ 
\XSolidBrush              & \CheckmarkBold         & \XSolidBrush           &  92.94         & 95.01  & 68.99 \\
\CheckmarkBold               & \XSolidBrush         & \XSolidBrush             & 93.88           &  96.05   & 69.37       \\ 
\XSolidBrush             & \CheckmarkBold           & \CheckmarkBold             & 94.03         & 97.82     & 69.88      \\ 
\CheckmarkBold               & \XSolidBrush          & \CheckmarkBold           & 94.51           & 98.90    & 70.44       \\ 
\CheckmarkBold             & \CheckmarkBold         & \XSolidBrush             & 95.38           & 99.37  & 71.54         \\ 
\CheckmarkBold             & \CheckmarkBold         & \CheckmarkBold           & \textbf{96.93}   & \textbf{99.92} & \textbf{72.32}   \\ \hline
\end{tabular}}
\caption{Ablation study of \emph{CorrDetail} with FF++ datasets (LQ and HQ). Best results are highlighted in \textbf{bold}. All scores are in \%.
%AUC (↑) indicates higher values are better. Best results are highlighted in red. All metric units are expressed in percentages (\%).
}
\label{tab:ablation_single_auc}
\end{table}

\subsection{Ablation Study}
In this section, we discuss practical effectiveness of each module in \emph{CorrDetail}, including the enhancement of learning in SCVQA compared to VQA, the effectiveness of CFDE in accurately enhancing target semantics within images, and, finally, whether the decision enhancement module can effectively address certain challenging samples that VLMs struggle to handle. Detailed results are listed in Table~\ref{tab:ablation_single_auc}.

\textbf{Illustrating the Effectiveness of SCVQA.} 
To demonstrate the performance of SCVQA, we compared models trained using SCVQA with the GPT-4o mini and visualized the outputs on several face-forged images, as shown in Fig.~\ref{fig:example1}. It is worth noting that during this ablation process, no additional innovative modules were incorporated—only prompt guidance was added. Clearly, for some images that 4o mini fails to classify correctly, our model accurately identifies the reasonable or unrealistic regions and details.

\begin{figure}[!t]
    \centering
    \includegraphics[width=0.48\textwidth]{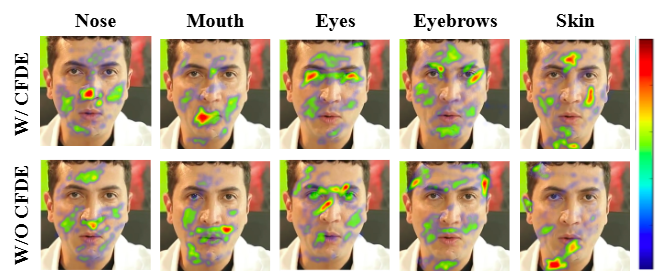}
    \caption{
    %The heatmap visualization of cross-modal semantic enhancement explicitly demonstrates the effectiveness of attention enhancement in this module.
    \fzl{
    The heatmap visualization of ablation study on CFDE.
    }
    }
    \label{fig:CSE_notice}
\end{figure}

\textbf{Discussing the influence of CFDE.} We compared the traditional prompt-based attention enhancement method with CFDE through heatmap visualizations, as shown in Fig.~\ref{fig:CSE_notice}. It is evident that feature-level processing yields a more focused overall effect compared to traditional methods. For instance, at the \textit{skin} level, our method produces a broader heat distribution compared to the prompt-based method, while at the \textit{nose} level, it exhibits more concentrated heat around the corresponding parts of the image. This comprehensive comparison demonstrates that enhancing from a cross-modal semantic perspective is more effective in guiding the model's recognition than relying solely on text-based approaches.

\textbf{Discussing the influence of decision fusion.} In this section, we focus on addressing native negative sample issues that can be resolved through integrated decision-making, as shown in Fig.~\ref{fig:decision fuse}. For certain images where the background occupies a large proportion of the frame or where the face is slightly blurred, the VLM's attention to the facial region becomes weaker, leading to incorrect classifications. By visualizing the outputs, we further demonstrate the effectiveness of post-processing. Meanwhile, the enhancements from the visual module are more reflected in the overall improvement of classification performance, as shown in Table~\ref{tab:ablation_single_auc}.

\section{Conclusion}

In this paper, we present \emph{CorrDetail}, a novel face forgery detection framework that enhances both performance and interpretability through its key modules. The self-correction mechanism utilizes error-guided questions to improve the VLM’s efficiency and generalization, achieving SOTA results in face forgery detection tasks. Our visual fine-grained detail enhancement module provides precise forgery cues by strengthening specific semantic features. Additionally, the fused decision strategy employs post-processing to address focus issues the model cannot resolve independently. \emph{CorrDetail} outperforms recent methods across multiple benchmarks, demonstrating superior accuracy and generalization.

\section*{Acknowledgments}
% no 序号

This work is funded by National Key Research and Development Project (Grant No: 2022YFB2703100),  Ant Group through CCF-Ant Research Fund, and Information Technology Center, ZheJiang University.
\section*{Contribution Statement}

Authors with † have made the same contribution to this work.

\bibliographystyle{named}
\bibliography{ijcai19}

\end{document}